# Limitations of Skeptical Default Reasoning

Jens Doerpmund
Department of Computer Science
University of Manchester,
Manchester M13 9PL, UK
jens@cs.man.ac.uk

## Abstract

Poole has shown that nonmonotonic logics do not handle the lottery paradox correctly. In this paper we will show that Pollock's theory of defeasible reasoning fails for the same reason: defeasible reasoning is incompatible with the skeptical notion of derivability.

## 1 Introduction

In the preface to "Cognitive Carpentry: A Blueprint for How to Build a Person" John Pollock argues that "work in artificial intelligence has made less progress than expected because it has been based upon inadequate foundations. AI systems are supposed to behave rationally, but their creators have not generally come to the problem equipped with an account of what it is to behave rationally." OSCAR, Pollock's framework for a reasoning engine, on the other hand, is "capable of performing reasoning that philosophers would regard as epistemologically sophisticated [Pollock, 1995]."

Pollock claims that because many theories of uncertain reasoning are semantical (i.e. proceeding in terms of models) rather than syntactical (i.e. proceeding in terms of arguments which may defeat one another), they are "unable to adequately formulate principles of probabilistic reasoning [Pollock, 1990, Chapter 8]." Unfortunately, the fact that Pollock's framework is not characterized semantically makes it very difficult to determine which conclusions it sanctions. It may be for this reason that the mistake in Pollock's analysis regarding the important distinction between the paradox of the preface and the lottery paradox ([Pollock, 1990], [Pollock, 1994], [Pollock, 1995]) remained undetected. This is very unfortunate since other relevant problems of default reasoning (i.e. preference for more specific knowledge and contrapositive reasoning) pose no problem in Pollock's framework. Let us start with a review of the lottery paradox in default reasoning.

## 2 Default Reasoning and the Lottery Paradox

In standard logic we are not able to draw conclusions if the available evidence is insufficient to guarantee their correctness. Default reasoning, on the other hand, allows us to jump to conclusions even if the knowledge about the domain at hand is incomplete. Some of these conclusions may turn out to be incorrect and must be abandoned in order to avoid inconsistencies. To make this more precise, we provide a brief introduction to Reiter's default logic [Reiter, 1980].

### 2.1 Default Logic

Reiter's default logic supports rules like 'if $a(x)$ is provable and if it is consistent to belief $b(x)$, then derive $c(x)$.' These rules are not expressed *in* the underlying object language, but can be seen as additional inference rules which augment those provided by first-order logic. Such a rule, which is called a *default*, has the following form:

$$\frac{a(\bar{x}) : b_1(\bar{x}), ..., b_n(\bar{x})}{c(\bar{x})} \quad (1)$$

where $a(\bar{x})$, $b_1(\bar{x})$, ..., $b_n(\bar{x})$, and $c(\bar{x})$ are first-order formulae whose free variables are among those of $\bar{x} = (x_1, ..., x_m)$. The meaning of a default is that $c(\bar{x})$ is believed if $a(\bar{x})$ is known to be true and $b_1(\bar{x})$, ..., $b_n(\bar{x})$ are consistent with what is known (i.e. $\neg b_i(\bar{x})$ cannot be deduced). A default is called *applicable* if both of these conditions are satisfied; otherwise it is *inapplicable*. For example the statement "Birds usually fly" is expressed by the default

$$\frac{bird(x) : flies(x)}{flies(x)}. \quad (2)$$

A *default theory* $\Delta$ is a pair $(D, W)$ of sets, where $W$ contains ordinary first-order sentences representing knowledge about a particular domain, and $D$ is a set of defaults, which are used to (partially) extend this knowledge by making plausible, but logically invalid inferences. Belief sets which are obtained by taking all formulae which can be deductively derived from both $W$ and the consequents of all applicable defaults of $D$ are called *extensions* of the theory $\Delta$.[1]

## 2.2 Multiple Extensions

A problem of default logic is that we often obtain *multiple extensions*. Consider the following classical example:

**Example 1** (Nixon-Diamond)
*Quakers are typically pacifists.*
*Republicans are typically not pacifists.*
*Nixon is a Quaker and a Republican.*
*Is Nixon a pacifist?*

This can be encoded into the default theory $\Delta = (D, W)$, where $D =$

$$\left\{ \frac{Quaker(x) : pacifist(x)}{pacifist(x)}, \frac{Republican(x) : \neg pacifist(x)}{\neg pacifist(x)} \right\}$$

and $W = \{ Quaker(x) \land Republican(x) \}$.

The extensions of $\Delta$ are

$E_1 = $ Cn($\{Quaker(Nixon)$, $Republican(Nixon)$, $pacifist(Nixon)\}$) and
$E_2 = $ Cn($\{Quaker(Nixon)$, $Republican(Nixon)$, $\neg pacifist(Nixon)\}$),

The generation of multiple extensions is very common in logic-based frameworks for reasoning under uncertainty. What, then, can be inferred from the default theory of Example 1? According to Reiter, *pacifist(Nixon)* is provable, because each extension can be seen as an acceptable belief set and *pacifist(Nixon)* is contained in one of them. But by the same argument $\neg pacifist(Nixon)$ is provable. This "credulous" view which allows us to *choose* an extension does not seem to be a rational way of "proving" a sentence, since we can base our choice on what we actually want to infer. This problem is avoided in *skeptical default reasoning*, according to which a sentence is a theorem of a default theory if and only if the sentence is contained in (the intersection of) all extensions.[2] In Example 1, we would therefore not be justified in believing either *pacifist(Nixon)* or $\neg pacifist(Nixon)$. Although this seems to be the correct approach in this particular example, we will soon see that there are default theories which lead to rather peculiar results.

## 2.3 The Lottery Paradox

Consider the following example taken from [Poole, 1991, p. 290]:

---

[1]See [Reiter, 1980] for a formal definition of an extension.

[2]Note that the intersection of extensions is itself not an extension, since there are always applicable defaults which have not been applied.

**Example 2** (Qualitative Lottery Paradox) *Suppose we want to build a knowledge base about birds. Suppose also that all we are told about Tweety is that Tweety is a bird. We first state knowledge about the different birds we are considering:*

$\forall x. bird(x) \equiv emu(x) \lor penguin(x) \lor hummingbird(x)$
$\lor sandpiper(x) \lor albatross(x) \lor ... \lor canary(x)$.

*We now state defaults about birds (e.g., they fly, are within certain size ranges, nest in trees, etc.). For each sort of bird that is exceptional in some way we will be able to conclude that Tweety is not that sort of bird:*

- *We conclude that Tweety is not an emu or a penguin because they are exceptional in not flying.*
- *We conclude that Tweety is not a hummingbird as hummingbirds are exceptional in their size [...]*

*The reason that we divide the class of birds into subclasses is because each subclass is exceptional in some way. Rather than being a pathological example, this would seem to be the general rule, typical of hierarchies with exceptions.*

This example is a qualitative version of the lottery paradox [Kyburg, Jr., 1961]. The original version is as follows:

**Example 3** (Lottery Paradox) *Consider a fair lottery consisting of one million tickets, only one of them is a winning ticket. Suppose you possess one ticket. Since the probability of winning the prize is 0.000001, it seems reasonable to infer that your ticket will not win. By the same argument it is reasonable to infer that* each *ticket will not win. This, however, is inconsistent since one ticket must be the winning ticket.*

In both of the above-mentioned examples we end up having inconsistent beliefs. In default logic, however, we are only allowed to apply defaults if it is consistent to do so. Thus, in the lottery we would infer that the first 999,999 tickets will lose and – in order to maintain consistency – that ticket No. 1,000,000 will win. Since the order in which defaults are applied is irrelevant, we will get 999,999 other extensions. The intersection of all extensions does not contain any default inferences whatsoever. Thus, although it is very likely that, say, ticket No. 4711 is a losing ticket, we are not able to infer this in default logic. In the qualitative example, we might infer in one extension that Tweety is an emu, in another extension that Tweety is a sand-piper. The number of extensions will be the same as the number of different kinds of birds. Since there are extensions in which Tweety is a non-flying bird, we are not able to infer that Tweety can fly if we opt for the skeptical view and decide to accept only beliefs contained in the intersection of all extensions.

The lottery paradox is indeed a serious problem in default reasoning. It arises whenever jointly inconsistent defaults are equally strong. Often, however, one default should be preferred over another default. This is normally the case when one default is *more specific* than a competing one. Unfortunately, default reasoning systems, as described in [Poole, 1984] or [Reiter, 1980] are not capable of deciding which defaults are more specific and should therefore be preferred (i.e. applied first). Contraposition is yet another problem in default reasoning. For instance, knowing that birds normally fly, we may want to infer that an individual that can not fly is (normally) not a bird. But often contrapositive reasoning is not desired. We certainly don't want to infer that something that can fly is not an object, given that objects normally don't fly. Again, default reasoning systems are not capable of deciding under which circumstances the contraposition on defaults should be considered.

## 3 Pollock's Theory of Defeasible Reasoning

Pollock's theory of defeasible reasoning constitutes a probabilistic framework that supports the detachment of beliefs by means of acceptance rules. Deductions are performed both on the level of probabilities and on the level of beliefs. The simplest acceptance rule, called (A1), is as follows [Pollock, 1990].

**Definition 1 (A1)** *If $F$ is projectible[3] with respect to $G$ and $r > 0.5$ then $G(c) \wedge \Pr(F \mid G) \geq r$ is a prima facie reason for $F(c)$, the strength of the reason depending upon the value of $r$.*

The reasons for beliefs we obtain by applying (A1) are only prima facie reasons – that is, they can be defeated by so-called defeaters. Pollock distinguishes between *undercutting defeaters*, which attack the connection between a prima facie reason and its conclusion, and *rebutting defeaters*, which are reasons for denying a conclusion. A belief is *warranted* if and only if it has at least one prima facie reason which cannot be defeated.

An important undercutting defeater is the following.

**Definition 2 (D1)** *If $F$ is projectible with respect to $H$, then $H(c) \wedge$
$\Pr(F \mid G \wedge H) \neq \Pr(F \mid G)$ is an undercutting defeater for (A1).*

(D1) defeats prima facie reasons generated by (A1) whenever (A1) does not take all relevant information into account. Thus, the preference of reasons based on more specific information is dealt with correctly and does not need to be provided as in default logic.

An fundamental principle of Pollock's framework is the principle of collective defeat [Pollock, 1990, pp. 88].

---

[3]The projectibility constraint is explained in [Pollock, 1990]. It is of no importance to our discussion.

**Definition 3 (Collective Defeat)** *If we have equally good independent reasons for believing each member of a minimal set of jointly inconsistent propositions, and none of the reasons is contradictory in any other way, then none of the propositions in the set is justified.*

The justification for the priciple of collective defeat is easily stated. Each $P_i$ of a minimal set of inconsistent propositions $\{P_1, ..., P_N\}$ is rebutted (and therefore not warranted), because its negation is deductively entailed by the conjunction $P_1 \wedge ... \wedge P_{i-1} \wedge P_{i+1} \wedge ... \wedge P_N$.

Note that the principle of collective defeat is very similar to the skeptical notion of derivability in default logic. In fact, in the lottery example Pollock's framework sanctions the same conclusions as we would get in default logic. For instance, in Example 2 collective defeat forces us to remain agnostic about what kind of bird Tweety is. But this, it seems, does not affect the inference from the evidence stating that most birds fly to our believing that Tweety can fly.

However, in "The Collapse of Collective Defeat: Lessons from the Lottery Paradox," Kevin Korb constructs an argument against collective defeat that does not rely on the generation of multiple extensions. His criticism is based on the fact that "one can 'lotterize' just about any inductive inference problem, and so, if using Pollock's Rule [i.e. collective defeat], one will *almost always* be constrained to indecision, even concerning the most ordinary, dull, unobjectionable inferences" [Korb, 1992]. Lotterization, of course, refers to the process of dividing the sample space of an experiment into many partitions such that each member of the partition has an equally low probability of being selected. Korb's argument, then, is as follows. Let $P$ be a proposition that has a high probability, and let Q be the negation of $P$. Normally, this would give us a reason for concluding $P$ (and consequently $\neg Q$). We now lotterize both $P$ and $Q$ such that all members of $P$ and $Q$ have roughly the same low probability. Thus, our knowledge base contains the following information:

$$P \equiv \neg Q,$$
$$P \equiv P_1 \vee P_2 \vee ... P_M,$$
$$Q \equiv Q_1 \vee Q_2 \vee ... Q_N,$$
$$\Pr(P) = 1 - \Pr(Q) = \text{'high'},$$
$$\Pr(\neg P_i) \approx \Pr(\neg Q_j) = \text{'high'},$$
$$\text{for all } i \in \{1,..,M\}, j \in \{1,...,N\}.$$

Since exactly one disjunct of

$$P_1 \vee ... \vee P_M \vee Q_1 \vee ... \vee Q_N$$

is true, collective defeat applies and prevents us from believing the negations of any of the disjuncts. In particular, we are not warranted in believing any of the $\neg Q_i$. Our reasoning so far has had no influence on the high probability of $P$. Nevertheless, we cannot conclude $P$ anymore. Suppose $P$ were warranted – that is, we have no reason for believing $\neg P$ that is as strong as the reason for believing $P$. Then we would be able to infer any of the $\neg Q_i$ deductively. But, as we have seen, collective defeat prevents us from concluding any of the $\neg Q_i$. Therefore, Korb argues, P cannot be warranted.

But Korb's argument is wrong. Precisely because we can infer $\neg Q_i$ from $P$, the partition Korb has chosen is not minimal any more.[4] The minimal partition would just contain the $\neg P_i$. In other words, we would not be justified in believing any of the $\neg P_i$ (because of collective defeat), but we would be able to infer each of the $\neg Q_i$, because $P$ is warranted. If, on the other hand, $P$ is not warranted, then Korb's argument is correct but irrelevant, because we do not need further arguments against $P$.

### 3.1 The Paradox of the Preface vs. the Lottery Paradox

Even proponents of collective defeat acknowledge that there are situations in which the principle should not be applied. For instance, a person (vacuously) believes all of his beliefs, but he also believes that some of his beliefs are wrong (although he does not know which). Should he apply the principle of collective defeat and remain agnostic about everything he believes? Consider the following problem taken from [Makinson, 1965].

> Suppose that in the course of his book a writer makes a great many assertions, which we shall call $s_1, ..., s_n$. Given each one of these, he believes that it is true. If he has already written other books, and received corrections from readers and reviewers, he may also believe that not everything he has written in his latest book is true. His approach is eminently rational; he has learnt from experience. The discovery of errors among statements which previously he believed to be true gives him a good ground for believing that there are undetected errors in his latest book [Makinson, 1965].

The structure of this *paradox of the preface*[5] is very similar to the lottery paradox. As in the lottery case, we seem to be justified in believing each statement $s_1, ..., s_n$ while at the same time we are justified in believing $\neg s_1 \vee ... \vee \neg s_n$. However, if we applied the principle of collective defeat to the paradox of the preface, we would no longer be justified in believing any of the statements mentioned in the book! Examples like the paradox of the preface arise naturally in default reasoning. What distinguishes default reasoning from deductive reasoning is that the inferences in default reasoning need not be valid, but merely plausible. Suppose, then, that the conclusions we derive from our evidence have a probability of 99.9% on average. Given that we have made more then 1000 inferences, we should have a good reason for believing that at least one of the conclusions is wrong. If we were to apply collective defeat in this case, we would not be able to perform default reasoning at all.

Pollock therefore provides a reason for treating instances of the lottery paradox and the paradox of the preface differently. The tickets in the lottery, Pollock argues, are *negatively relevant*. That is, inferring that a ticket loses makes it more likely that one of the remaining tickets is the winning ticket. We therefore cannot possibly conclude that all tickets are winning tickets without obtaining an inconsistent belief set. Contrast this with the paradox of the preface. Here, the statements made in the book are *positively relevant*. Confirming one statement decreases the probability that there exists a wrong statement, which in turn makes the other statements more likely [Pollock, 1995, p. 128]. We could, therefore, consistently believe each of the statements, and, as a consequence, discard the belief stating that (at least) one statement is wrong.

Thus, rather than seeing the above examples as arguments supporting the view that uncertain reasoning frameworks be able to handle inconsistent information, Pollock prefers to ignore that the belief set is probably inconsistent. This view is not uncommon. In [Lehrer, 1990], for instance, it is argued that if we accept that at least one of our beliefs is wrong, then we forgo the chances of achieving the ideal of *maxiverificity*. A person is called *maxiverific* if he accepts all true statements and rejects all false statements by some rational methodology. Lehrer claims that it is not irrational for a sensible person to "adopt maxiverificity as an ideal and refuse to follow a policy that guarantees failure [Lehrer, 1990]." Lehrer therefore comes to the same conclusion as Pollock: he suggests to ignore that some of the beliefs may be wrong.[6]

Let us restate Pollock's argument regarding positive and negative relevance in more detail. We know that most statements contained in a book are true, but we also know that it is very likely that at least one of them is false. We therefore have prima facie reasons for believing each of the statements $s_i$ as well as

$$\neg s_1 \vee ... \vee \neg s_N. \qquad (3)$$

These beliefs are inconsistent, and it appears that they are subject to collective defeat. Although we would like to believe each of the $s_i$, we can construct arguments for their negations. Suppose the statements $s_1, ..., s_{j-1}, s_{j+1}, ..., s_N$ are warranted. After all, each statement has a high probability of being true. From

---

[4]Cf. Definition 3 for a definition of collective defeat.

[5]The name of the paradox refers to the preface of a book in which it is common that authors acknowledge the support of various people, but take responsibility for any mistakes remaining in the book.

[6]However, Lehrer acknowledges that there are other strategies that may be equally rational.

this we can conclude

$$s_1 \wedge ... \wedge s_{j-1} \wedge s_{j+1} \wedge ... \wedge s_N. \tag{4}$$

(3) and (4) together deductively imply $\neg s_j$, and it seems that we are no longer justified in believing $s_j$. However, we can show that the argument for $\neg s_j$ is defeated. Since the statements in the book are *not negatively relevant* to each other, the confirmation of some statements does not make the remaining statements more likely. Instead, confirming a statement decreases the probability that the book contains false statements.

**Claim:**

The following equation, which, according to (D1), provides an undercutting defeater for the inference leading to (3), holds.

$$\begin{aligned} &\Pr(\neg s_1 \vee ... \vee \neg s_N \mid \\ &\quad s_1 \wedge ... \wedge s_{i-1} \wedge s_{i+1} \wedge ... \wedge s_N) \\ < &\quad \Pr(\neg s_1 \vee ... \vee \neg s_N) \end{aligned} \tag{5}$$

**Proof:**

If $\{s_1, ..., s_N\}$ is a set of propositions that are positively relevant, then

$$\Pr(s_i \mid s_1 \wedge ... \wedge s_{i-1} \wedge s_{i+1} \wedge ... \wedge s_N) > \Pr(s_i). \tag{6}$$

Since

$$\begin{aligned} &\Pr(s_i \mid s_1 \wedge ... \wedge s_{i-1} \wedge s_{i+1} \wedge ... \wedge s_N) \\ = &\quad 1 - \Pr(\neg s_i \mid s_1 \wedge ... \wedge s_{i-1} \wedge s_{i+1} \wedge ... \wedge s_N) \\ = &\quad 1 - \Pr(\neg s_1 \vee ... \vee \neg s_N \mid \\ &\quad s_1 \wedge ... \wedge s_{i-1} \wedge s_{i+1} \wedge ... \wedge s_N) \\ > &\quad \Pr(s_i) = 1 - \Pr(\neg s_i), \end{aligned} \tag{7}$$

it follows that

$$\begin{aligned} &\Pr(\neg s_1 \vee ... \vee \neg s_N \mid \\ &\quad s_1 \wedge ... \wedge s_{i_1} \wedge s_{i+1} \wedge ... \wedge s_N) \mid \\ < &\quad \Pr(\neg s_i). \end{aligned} \tag{8}$$

Because the right hand side of (8) is less or equal than $\Pr(\neg s_1 \vee ... \vee \neg s_N)$, (5) holds. □

Therefore, (3) is not warranted, and we can conclude that all statements in the book are true.

According to Pollock, collective defeat applies if and only if the members of a minimal set of jointly inconsistent beliefs are negatively relevant [Pollock, 1994]. Clearly, if (3) were certain, then the statements would be negatively relevant. (3) states that (at least) one statement is false. Since all statements have the same probability of being true, the confirmation of one statement makes the other statements less likely if we also know that there are false statements. However – and this is where Pollock is mistaken, the converse is not true. Collective defeat can be defeated even if the propositions are negatively relevant.

Consider a lottery which is not always fair: every 100th draw does not contain a winning ticket. Since most draws are fair, we have a prima facie reason for believing $t_1 \vee ... \vee t_{1000000}$. Clearly, confirming that some tickets are losing tickets makes it more likely that the winning ticket is among the remaining tickets. Thus, the tickets are negatively relevant. However, the more tickets we confirm, the more probable it is that the current draw contains only losing tickets. Thus, we have an undercutting defeater for concluding

$$t_1 \vee t_2 \vee ... \vee t_{1000000} \tag{9}$$

despite the fact that the tickets are negatively relevant:

$$\begin{aligned} &\Pr(t_1 \vee t_2 \vee ... \vee t_{1000000} \mid \\ &\quad \neg t_1 \wedge ... \wedge \neg t_{i-1} \wedge \neg t_{i+1} \wedge ... \wedge \neg t_{1000000}) \\ < &\quad \Pr(t_1 \vee t_2 \vee ... \vee t_{1000000}). \end{aligned} \tag{10}$$

Without (9) being warranted, we can infer that all tickets are losing tickets.

Because our argument does not depend on the negative relevance of the propositions in the belief set, it has a devastating consequence. We can basically lotterize any proposition $P$ having a high probability that is smaller than 1.

**Claim:**

Let $P \equiv p_1 \vee p_2 \vee ... \vee p_N$ be a 'lotterization' of $P$. If $N$ is sufficiently large, we have a reason for believing each of the $\neg p_i$, even when the $p_i$ are not positively relevant. This, however, means that $P$ can never be warranted.

**Proof:**

We need to show that we are justified in believing that all $p_i$ are false. In order to do this we show that the following equation, which provides a defeater for believing $p_1 \vee ... \vee p_N$, holds.

$$\begin{aligned} &\Pr(p_1 \vee ... \vee p_N \mid \\ &\quad \neg p_1 \wedge ... \wedge \neg p_{i-1} \wedge \neg p_{i+1} \wedge ... \wedge \neg p_N) \\ < &\quad \Pr(p_1 \vee ... \vee p_N). \end{aligned} \tag{11}$$

Furthermore, we need to show that

$$\begin{aligned} &\Pr(p_1 \vee ... \vee p_N \mid \\ &\neg p_1 \wedge ... \wedge \neg p_{i-1} \wedge \neg p_{i+1} \wedge ... \wedge \neg p_N) \end{aligned} \tag{12}$$

is less than 0.5 in order to make sure that the acceptance rule cannot be applied to create an undefeated reason for believing $p_1 \vee ... \vee p_N$.

The probability of $P$ being false given that all but one of the $p_i$ are false is

$$\Pr(\neg P \mid \neg p_1 \wedge ... \wedge \neg p_{j-1} \wedge \neg p_{j+1} \wedge ... \wedge \neg p_N)$$

$$= \frac{\Pr(\neg P \wedge \neg p_1 \wedge ... \wedge \neg p_{j-1} \wedge \neg p_{j+1} \wedge ... \wedge \neg p_N)}{\Pr(\neg p_1 \wedge ... \wedge \neg p_{j-1} \wedge \neg p_{j+1} \wedge ... \wedge \neg p_N)}$$

$$= \frac{\Pr(\neg P)}{\Pr(\neg p_1 \wedge ... \wedge \neg p_{j-1} \wedge \neg p_{j+1} \wedge ... \wedge \neg p_N)} \quad (13)$$

If we know that all but one of the $p_i$ are false, then either the remaining $p_j$ is true, or $P$ is false. Thus,

$$\Pr(p_j \vee \neg P \mid \neg p_1 \wedge ... \wedge \neg p_{j-1} \wedge \neg p_{j+1} \wedge ... \wedge \neg p_N) = 1. \quad (14)$$

It follows that

$$\Pr(\neg p_1 \wedge ... \wedge \neg p_{j-1} \wedge \neg p_{j+1} \wedge ... \wedge \neg p_N)$$
$$= \Pr(\neg P) + \Pr(p_j \mid P)$$
$$= \Pr(\neg P) + \frac{1}{N}. \quad (15)$$

From (13), (14) and (15) we obtain

$$Pr(p_j \mid \neg p_1 \wedge ... \wedge \neg p_{j-1} \wedge \neg p_{j+1} \wedge ... \wedge \neg p_N)$$
$$= 1 - Pr(\neg p_j \mid \neg p_1 \wedge ... \wedge \neg p_{j-1} \wedge \neg p_{j+1} \wedge ... \wedge \neg p_N)$$
$$= 1 - \frac{\Pr(\neg P)}{\Pr(\neg p_1 \wedge ... \wedge \neg p_{j-1} \wedge \neg p_{j+1} \wedge ... \wedge \neg p_N)}$$
$$= 1 - \frac{\Pr(\neg P)}{\Pr(\neg P) + \frac{1}{N}}. \quad (16)$$

The right hand side of equation (16) decreases as $N$ increases. Thus, we can make

$$\Pr(p_j \mid \neg p_1 \wedge ... \wedge \neg p_{j-1} \wedge \neg p_{j+1} \wedge ... \wedge \neg p_N) \quad (17)$$

as small as we want. □

$P$ is no longer warranted if the value of (16) is smaller then 0.5, or equivalently, if

$$N > \left\lceil \frac{1}{\Pr(\neg P)} \right\rceil. \quad (18)$$

## 4 Conclusions

We have shown that any proposition $P$ which has a probability of less than 1.0 is not accepted in Pollock's framework. The reason is not – as Kolb believes – because we could lotterize $P$ and then show that the obtained disjuncts undergo collective defeat. Instead, the problem is that for each $P \equiv p_1 \vee ... \vee p_N$, we can create a defeater

$$\Pr(p_1 \vee ... \vee p_N \mid$$
$$\neg p_1 \wedge ... \wedge \neg p_{j-1} \wedge \neg p_{j+1} \wedge ... \wedge \neg p_N)$$
$$< \Pr(p_1 \vee ... \vee p_N). \quad (19)$$

But does the left-hand side of the above equation really provide more information than the right-hand side?

Limitations of Skeptical Default Reasoning    155

Clearly, each of the $\neg p_i$ has a high probability and is therefore justified. However, given that one of the $p_i$ must be true, and that we don't know which one it is,

$$Pr(\neg p_1 \wedge ... \wedge \neg p_{j-1} \wedge \neg p_{j+1} \wedge ... \wedge \neg p_N)$$

is fairly low (namely $1/N$). Thus, the conjunction

$$\neg p_1 \wedge ... \wedge \neg p_{j-1} \wedge \neg p_{j+1} \wedge ... \wedge \neg p_N$$

should not be accepted and should therefore not contribute to provide a defeater for $P$. If Pollock's framework could be changed to take this into account, likely propositions could be warranted after all. But then we would no longer be able to distinguish between the lottery paradox and the paradox of the preface (and therefore not be justified in believing any statement contained in a book). In either case, Pollock's claim that OSCAR is "capable of performing reasoning that philosophers would regard as epistemologically sophisticated" does not withstand scrutiny.

## Acknowledgements

I would like to thank Ian Pratt for many valuable discussions which led to an earlier draft of this paper. Thanks also to the referees for their useful comments.

## References

[Korb, 1992] Korb, K. B. (1992). The collapse of collective defeat: Lessons from the lottery paradox. In *Proceedings of the Biennal Meetings of the Philosophy of Science Association*, volume 1, pages 230–236.

[Kyburg, Jr., 1961] Kyburg, Jr., H. E. (1961). *Probability and the Logic of Rational Belief*. Wesleyan University Press, Middletown, CT.

[Lehrer, 1990] Lehrer, K. (1990). *Metamind*, chapter 6: Reason and Consistency. Oxford University Press, New York.

[Makinson, 1965] Makinson, D. C. (1965). The paradox of the preface. *Analysis*, 25:205–207.

[Pollock, 1990] Pollock, J. L. (1990). *Nomic Probabilities and the Foundations of Induction*. Oxford University Press, New York.

[Pollock, 1994] Pollock, J. L. (1994). Justification and defeat. *Artificial Intelligence*, 67:377–407.

[Pollock, 1995] Pollock, J. L. (1995). *Cognitive Carpentry: A Blueprint for How to Build a Person*. MIT Press, Cambridge, Massachusetts.

[Poole, 1984] Poole, D. L. (1984). A logical system for default reasoning. In *Proc. of the Non-Monotonic Reasoning Workshop*, pages 373–384, New Paltz, NY.

[Poole, 1991] Poole, D. L. (1991). The effect of knowledge on belief: conditioning, specificity and the lottery paradox in default reasoning. *Artificial Intelligence*, 49:281–307.

[Reiter, 1980] Reiter, R. (1980). A logic for default reasoning. *Artificial Intelligence*, 13:81–132.